\documentclass[usletter, 10pt, conference]{IEEEtran}

\usepackage[utf8]{inputenc}
\usepackage[T1]{fontenc}
\usepackage{times}
\usepackage{graphicx}
\usepackage{amsmath,amssymb}
\usepackage{booktabs}
\usepackage{siunitx}
\usepackage{url}
\usepackage{balance}
\usepackage{hyperref}

\newcommand{\link}[2]{\noindent\textbf{#1:} \url{#2}\par}
\newenvironment{links}{\par}{\par}

\renewcommand{\baselinestretch}{0.985}
\usepackage{threeparttable}

\usepackage{xcolor}

\newcommand{\xmark}{\textcolor{red!70!black}{\ensuremath{\times}}}

\usepackage{cleveref}
\Crefname{equation}{Eq.}{Eqs.} 
\Crefname{section}{Sec.}{Secs.}
\Crefname{figure}{Fig.}{Fig.}
\Crefname{table}{Tab.}{tabs.}

\IEEEoverridecommandlockouts   
\title{\LARGE \bf MILER: Semantic Mid-Level Representation for Sim-to-Real Reinforcement Learning in Unstructured Autonomous Driving}
\author{
Thomas Steinecker$^{*}$, Denis Trescher$^{*}$, Alexander Bienemann, Thorsten Luettel, and Mirko Maehlisch \\
{\normalsize Chair of Machine Perception for Autonomous Driving, University of the Bundeswehr Munich, Germany} \\
\texttt{\{Thomas.Steinecker, Denis.Trescher\}@unibw.de}
\thanks{$^{*}$These authors contributed equally to this work.}
\thanks{This research paper is funded by dtec.bw – Digitalization and Technology Research Center of the Bundeswehr [project MORE] and the Federal Office of Bundeswehr Equipment, Information Technology and In-Service Support (BAAINBw) which we gratefully acknowledge. dtec.bw is funded by the European Union – NextGenerationEU.}%
}

\begin{document}
\maketitle
\thispagestyle{empty}
\pagestyle{empty}


\begin{abstract}

Reinforcement learning constitutes a promising approach owing to its potential for superhuman performance and self-learned policies. However, its application to real-world autonomous driving remains scarce, particularly in unstructured environments, because of the challenges associated with sim-to-real transfer for unstructured environments. In this work, we present MILER, an end-to-end policy framework with zero-shot sim-to-real transfer. During offline training, we employ a custom semantic mid-level representation (MLR) simulator and train the policy network using reinforcement learning, with its control outputs applied directly to a bicycle model. During deployment on the real vehicle, camera and LiDAR data are processed by BEVFusion to generate a semantic bird's-eye-view representation consistent with that of the MLR simulator. The actions generated by the policy network are not applied directly to the real vehicle. Instead, we employ a trajectory-alignment strategy that enables zero-shot sim-to-real transfer of both perception and control. We extensively evaluate the proposed framework on a diverse test track  comprising numerous challenges, including various obstacles, hairpin curves, velocities of up to \SI{33.6}{\kilo\meter\per\hour}, and off-road sections. In total, we drove \SI{17.3}{\kilo\meter} with two different vehicles on a \SI{3.0}{\kilo\meter} test track without human intervention, thereby demonstrating the effectiveness of our approach. Furthermore, the entire software stack runs on a Jetson AGX Orin.
\end{abstract}

\vspace{0.2cm}

\begin{links}
    \link{Evaluation video}{https://www.youtube.com/watch?v=IZli3Z87URI}
\end{links}

\section{Introduction}

Reinforcement learning (RL) has become increasingly popular across numerous fields~\cite{jumper2021highly, kaufmann2023champion, achiam2023gpt} owing to its potential to achieve superhuman performance through trial-and-error interactions with the environment. However, in real-world applications such as off-road autonomous driving, trial-and-error learning is unacceptable because of safety concerns. Consequently, RL agents are commonly trained in simulation. Simulations are, by nature, abstractions of reality, and the domain gap between simulated and real environments remains challenging and constitutes an open research question. This challenge is particularly pronounced in unstructured environments, which exhibit substantial diversity in both appearance and physical properties \cite{borges2022survey}.

\begin{figure}[t!]
    \centering
    \includegraphics[width=\columnwidth]{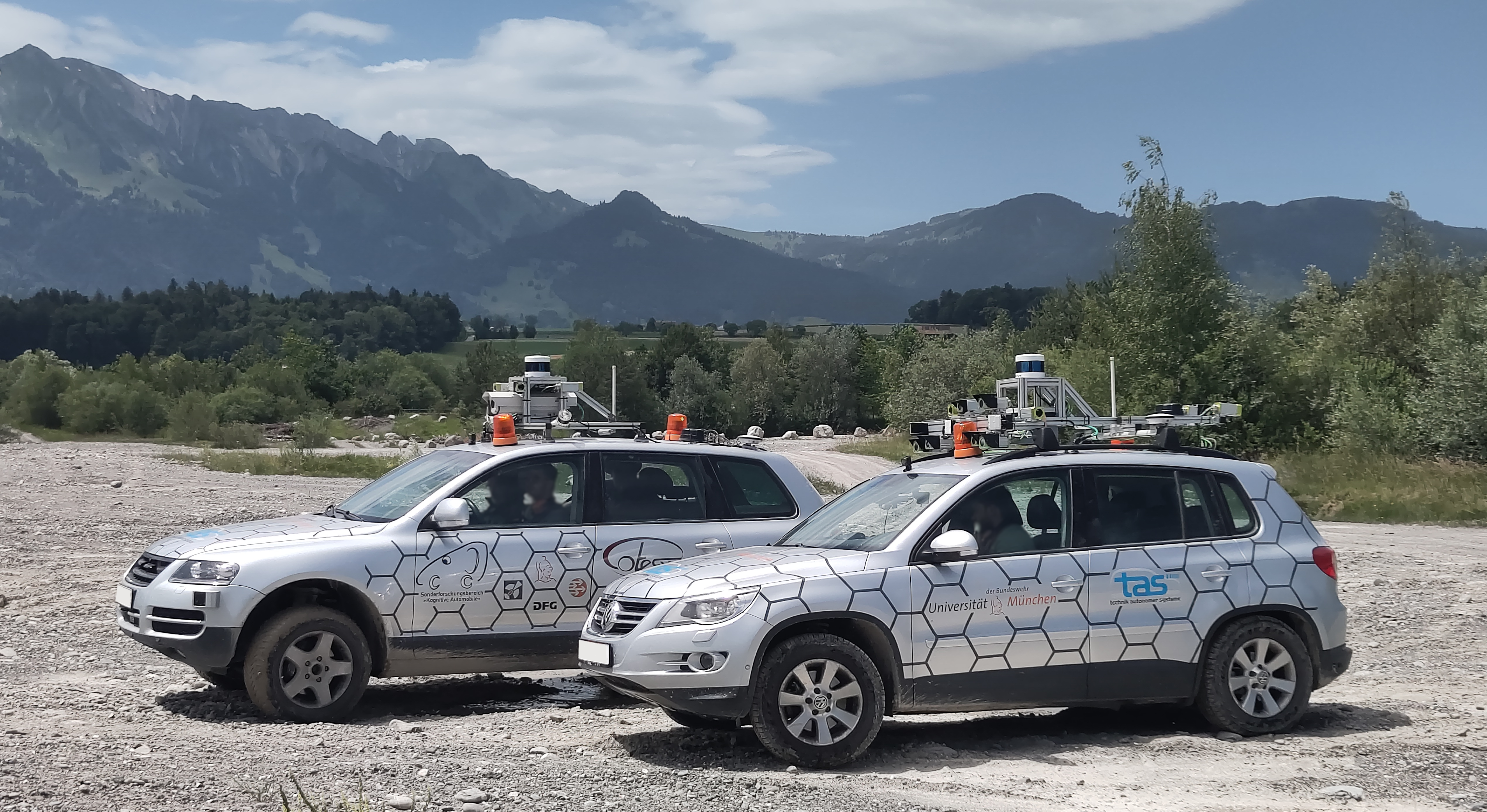}
    \caption{The two test vehicles (MuCAR-3, left; MuCAR-4, right) used for the real-world evaluation.}
    \label{fig:vehicles}
\end{figure}

\begin{figure*}[th!]
    \centering
    \includegraphics[width=1.0\textwidth]{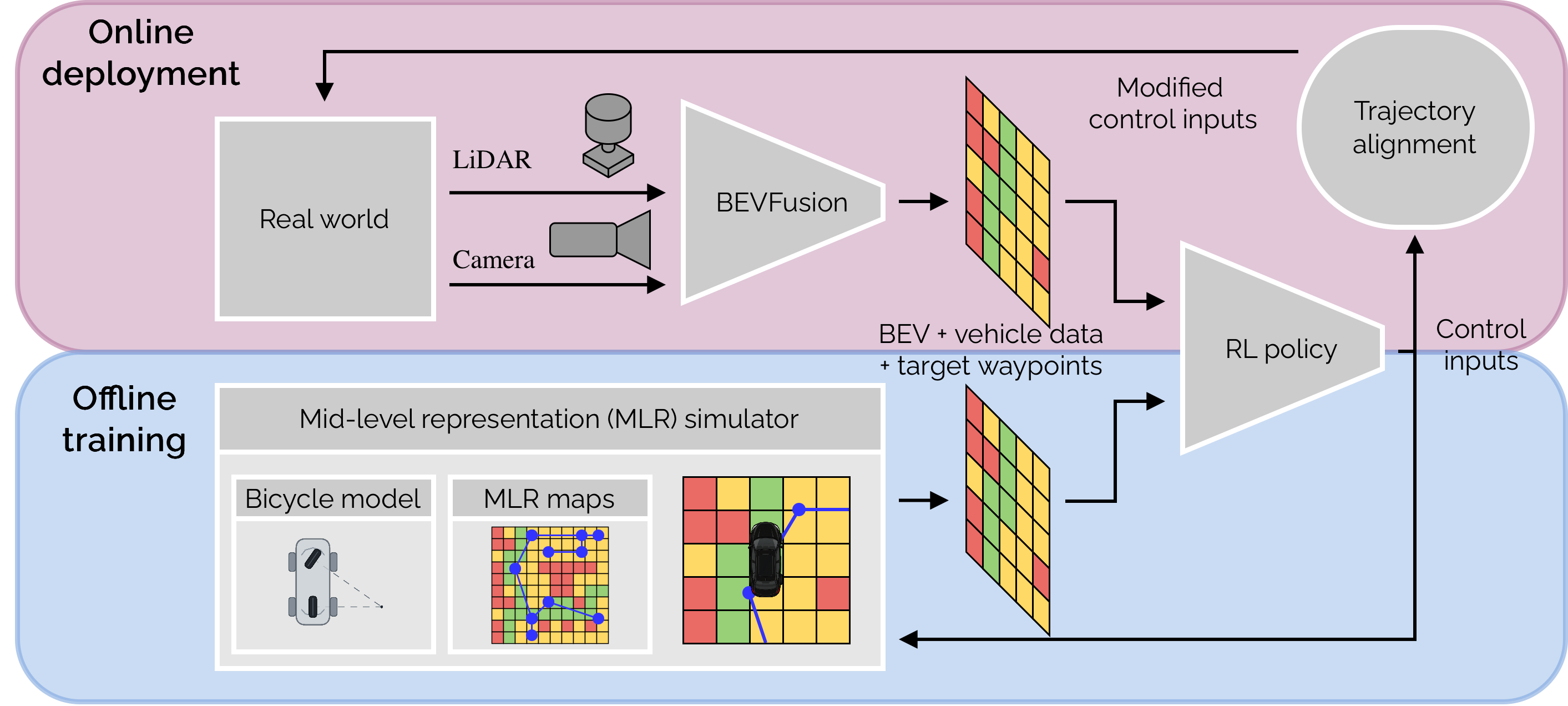}
    \caption{Simplified overview of the data-flow pipeline during offline training and online deployment. During training, a mid-level representation (MLR) simulator generates BEVs from semantic grid maps and extracts the vehicle state from the bicycle model, both of which are fed into the RL policy network. During online deployment, LiDAR and camera streams are processed by BEVFusion \cite{liu2023bevfusion} to generate semantic BEVs, which, together with the vehicle state measured by the vehicle sensors, are fed into the RL policy. Unlike in simulation, the actions are not applied directly to the real vehicle but are instead recalculated such that the real vehicle follows the simulated vehicle as closely as possible. 
    }
    \label{fig:overview}
\end{figure*}

In this work, we present an end-to-end RL-based approach for unstructured environments and provide a zero-shot sim-to-real transfer strategy, as illustrated in \Cref{fig:overview}. First, we train the RL agent in a semantic mid-level representation (MLR) simulator comprising synthetically generated maps from which bird's-eye-view (BEV) representations and target waypoints are extracted. The policy outputs are applied directly to the bicycle model employed by the simulator. For training we use Proximal Policy Optimization (PPO)~\cite{schulman2017proximal} including a discrete action head, curriculum learning, an undiscounted objective, and state normalization.

During deployment, we apply an adapted version of BEVFusion~\cite{liu2023bevfusion} to process camera and LiDAR data and generate BEVs that use the same representation as the MLR simulator. Applying the actions generated by the policy network directly to the real vehicle would result in unsafe driving behavior because of the considerable discrepancy between the bicycle model and the real vehicle arising from unmodelled time delays, friction, vehicle-model inaccuracies, and other effects. Instead, we build upon the trajectory-alignment strategy proposed in~\cite{steinecker2025dynamics}. The underlying idea is to simulate a virtual vehicle using the MLR simulator during deployment. This virtual vehicle receives the actions generated by the policy network, while the real vehicle is controlled longitudinally and laterally to minimize the pose difference between the two vehicles. 

Our main contributions are summarized as follows:
\begin{itemize}
    \item We introduce a semantic mid-level representation (MLR) simulator for unstructured environments that substantially reduces the required modelling effort.
    
    \item We introduce MILER, a zero-shot sim-to-real framework that enables the deployment of RL policies trained entirely in a highly abstract (bicycle model and MLR) simulation on two different real-world vehicles with little to no vehicle-specific adaptation.

    \item We provide an extensive evaluation of the proposed framework across many challenges on two different test vehicles (\Cref{fig:vehicles}) and demonstrate smooth driving at velocities of up to \SI{33.6}{\kilo\metre\per\hour} without notable oscillations. To the best of our knowledge, this is the first demonstration of a velocity above \SI{30}{\kilo\metre\per\hour} for a vehicle-sized, reinforcement-learning-based autonomous-driving application in an unstructured environment.
\end{itemize}

\subsection{Related Work}

\begin{table}[t!]
    \centering
    \caption{Comparison of end-to-end reinforcement learning approaches for off-road or unstructured autonomous driving. A dash indicates that the corresponding information was not reported.}
    \label{tab:offroad_comparison}
    \small
    \setlength{\tabcolsep}{1.0pt}
    \renewcommand{\arraystretch}{1.1}
    \begin{tabular}{lccccc}
        \toprule
        Method &
        Vehicle size &
        Max. speed &
        Simple sim. \\
        \midrule
        TADPO \cite{wu2026tadpo}
        & \checkmark
        & $\approx\SI{12.3}{\kilo\meter\per\hour}$%
        & \xmark \\

        WROOM \cite{kalaria2024wroom}
        & \xmark
        & —
        & \checkmark \\

        Sim2Seg \cite{so2022sim}
        & \checkmark
        & —
        & \xmark \\

        \textbf{Ours}
        & \checkmark
        & \textbf{\SI{33.6}{\kilo\meter\per\hour}}
        & \checkmark \\
        \bottomrule
    \end{tabular}
    \footnotetext{TADPO reports a mean real-world speed of
    \SI{3.41}{\meter\per\second}, corresponding to approximately
    \SI{12.3}{\kilo\meter\per\hour}, rather than an explicit maximum speed.}
\end{table}

The research for sim-to-real transfer for unstructured or off-road autonomous driving is scarce compared to the urban domain \cite{hu2023simulation}. 
Considerable work has investigated imitation learning~\cite{pan2017agile} and RL without simulation~\cite{kendall2019learning}. However, such approaches are unsuitable for safety-critical systems, for which simulators are required both to train policies and to evaluate their safety.

Sim-to-real transfer in robotic applications comprises two components: perception and actuation. Sophisticated actuation transfer is often neglected in applications involving low vehicle dynamics~\cite{folkers2019controlling}; however, discrepancies between simulated and real vehicle dynamics can cause instabilities as velocity increases. One prominent approach is to employ high-fidelity simulators~\cite{voogd2023reinforcement} that model vehicle dynamics and various interactions, including different ground types and height profiles as well as air resistance. A data-driven alternative is to learn the vehicle model through real-world interactions and subsequently use the learned model in simulation~\cite{maramotti2022tackling}. Another recent approach is trajectory alignment~\cite{steinecker2025dynamics}, which decouples the dynamics of the simulated and real vehicles and controls the real vehicle to follow the simulated vehicle by reducing their lateral and longitudinal deviations.

Similar to actuation transfer, sim-to-real transfer for perception is frequently addressed using high-fidelity simulators~\cite{wu2026tadpo, kalaria2024wroom}. Considerable effort is invested in modelling textures and physical lighting characteristics to reduce the domain gap for camera, LiDAR, and radar sensors. Sim2Seg~\cite{so2022sim} employs front-camera semantic segmentation as a mid-level representation; however, the approach still requires a high-fidelity simulator. \Cref{tab:offroad_comparison} provides a brief comparison with related approaches.

\begin{figure*}[th!]
    \centering
    \includegraphics[width=\textwidth]{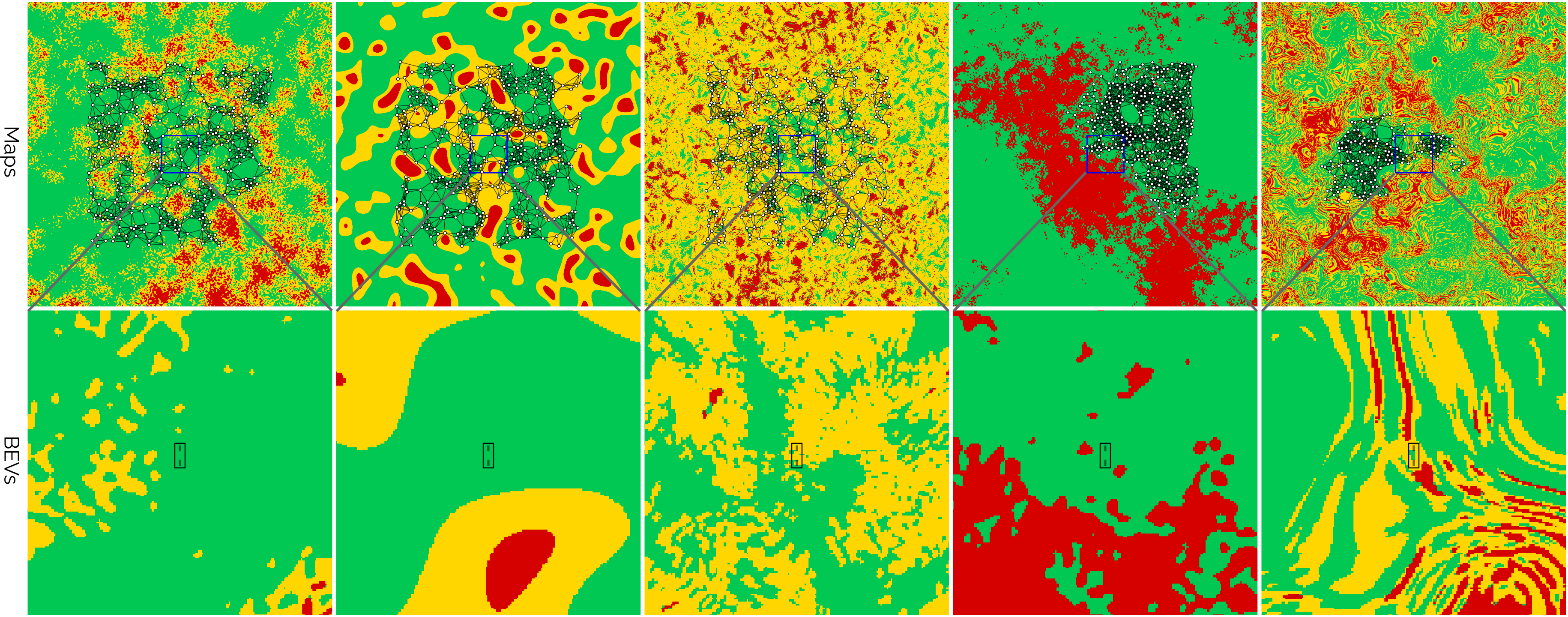}
    \caption{Examples of synthetically generated maps and corresponding
    BEVs. The top row shows five example maps with their target-waypoint
    networks. The bottom row shows BEVs obtained at the centers of the
    maps, including bounding boxes representing typical vehicle dimensions.
    The maps measure $\SI{500}{\metre} \times \SI{500}{\metre}$, and the
    BEVs measure $\SI{60}{\metre} \times \SI{60}{\metre}$.
    Classes: \texttt{Drivable}
    (\textcolor{green!60!black}{\rule{1.2ex}{1.2ex}}),
    \texttt{Semi-drivable}
    (\textcolor{yellow!80!orange}{\rule{1.2ex}{1.2ex}}), and
    \texttt{Non-drivable}
    (\textcolor{red}{\rule{1.2ex}{1.2ex}}).}
    \label{fig:maps_bevs}
\end{figure*}

Numerous simulators based on mid-level representations have been developed for structured urban driving. For example, Nocturne~\cite{nocturne2022} represents driving scenes using abstract road points, polylines, and traffic participants, whereas DriverGym~\cite{kothari2021drivergym} uses real-world driving data to construct rasterized BEV observations containing semantic elements such as roads and surrounding agents. Related abstract scene representations are employed by Waymax~\cite{gulino2023waymax}, CommonRoad~\cite{althoff2017commonroad}, and SMARTS~\cite{zhou2020smarts}. However, to the best of our knowledge, no mid-level representation simulator has previously been used for RL in unstructured autonomous-driving environments.


\section{Methods}

\subsection{Simulator}

We use a custom GPU-accelerated mid-level representation (MLR) simulator. The MLR comprises four semantic classes:
\begin{itemize}
    \item \texttt{Drivable}: paved roads and gravel;
    \item \texttt{Semi-drivable}: grass and tall grass; and
    \item \texttt{Non-drivable}: obstacles and stationary vehicles.
    \item \texttt{None}: unknown (required for the deployment strategy in \Cref{seq:deployment})
\end{itemize}
The advantage of an MLR simulator over realistic simulators such as CARLA~\cite{Dosovitskiy17} or MAVS~\cite{hudson2024mississippi} is the simplified generation of maps. Whereas the projection
\[
\substack{\text{3D environment}\\
          \text{(textures, geometry, \ldots)}}
\;\longrightarrow\;
\substack{\text{rendered sensor data}\\
          \text{(camera, LiDAR, \ldots)}}
\;\longrightarrow\;
\substack{\text{semantics}\\
          \text{(BEVs, voxels, \ldots)}}
\]
constitutes an infinite-to-one relationship, we entirely omit this projection from a computational perspective. Thus, neither the creation of realistic environments nor their rendering and projection are required during training. The vehicle motion is modelled using a bicycle model, while the vehicle geometry is represented by a rectangular bounding box.

\textbf{Map generation}—%
The map-generation process is inspired by Minecraft, which uses Perlin noise to generate height profiles. We overlay multiple layers with different grid sizes generated using Perlin noise and assign different value ranges to the semantic classes. This produces diverse maps with different characteristics, as shown in \Cref{fig:maps_bevs}. For training, we generate 3,800 maps with dimensions of $500\,\mathrm{m} \times 500\,\mathrm{m}$ and a resolution of $0.4\,\mathrm{m}$. For each map, we generate a connected graph with linear connections that is subsequently used to generate target waypoints. We randomly place graph nodes in drivable or semi-drivable areas while ensuring that they are sufficiently far apart. Nodes are connected if the linear connection between them is occupied by \texttt{Drivable} or \texttt{Semi-drivable} space.

\textbf{Episode generation}—%
In each episode, a random map is selected from a set of precomputed maps. Two random nodes are then selected from the graph and connected using breadth-first search. 
Subsequently, the resulting target waypoints are smoothed using the Curvature Corrected Moving Average (CCMA)~\cite{steinecker2023simple} to better align simulation and real characteristics, and it provides more informative learning feedback, since the heading can be used as reference, and improves robustness by matching training and real-world conditions. The target waypoints provide a rough indication of where to drive and cannot be followed without obstacle avoidance.
The initial node and its direction toward the subsequent node define the initial position and heading. However, the agent is spawned with Gaussian noise applied to its position, heading, velocity, acceleration, steering angle and steering rate to improve its robustness in the real-world application, where the agent will likewise not be placed perfectly on the target waypoints. Moreover, it promotes exploration by exposing the agent to situations that it might never encounter when following its own policy.

An episode is considered terminated if the agent remains stationary, collides with a non-drivable cell, or moves out of bounds, defined as a lateral deviation of more than $25\,\mathrm{m}$ from the target waypoints. An episode is considered finished only if the agent is stationary within $25\,\mathrm{m}$ of the final waypoint. Additionally, the simulator generates randomly changing target velocities that the agent should strive to achieve without exceeding them.

\textbf{BEV generation}—%
The map is a grid of cells, while the agent can move continuously across these cells. For each BEV cell center relative to the ego vehicle, the semantic class is assigned using the nearest neighbor based on the distance between the BEV cell center and the map cell centers. To prevent map cells from being temporarily unobservable, the following condition must hold for the cell width $w$:
\[
    w_{\mathrm{map}}
    \geq
    \sqrt{2}\,w_{\mathrm{BEV}}.
\]

\subsection{Modelling}
\label{sec:modelling}

\textbf{State modelling}—%
The agent state comprises six components:
\begin{itemize}
    \item The BEV, which contains the four semantic classes, is centered on the vehicle, has a total length of \SI{60}{\metre}, and a cell resolution of $\SI{25}{\centi\metre} \times \SI{25}{\centi\metre}$. The BEV is embedded using one-hot encoding. In addition, the target waypoints are rasterized into an additional channel, which places the route in the same spatial frame as the obstacles. We use a single BEV frame approach which avoids temporal inconsistencies for sim-to-real transfer.
    
    \item The vehicle state $\mathbf{s}_{\mathrm{veh}}$, consisting of the velocity, acceleration, steering angle, and steering rate. (Position and heading are redundant because of the ego-centered coordinate frame.)
    
    \item The previous actions $\mathbf{a}_{t-1}$, to account for smooth driving.
    
    \item The target velocity $v_{\text{target}}$.
    
    \item List of the next 80 waypoints $\mathbf{W}_t
=
\left[
    (x_{\mathrm{wp},i},y_{\mathrm{wp},i})
\right]_{i=1}^{N_{\mathrm{wp}}}$ in the ego frame, each comprising the position, spaced $\SI{1}{\metre}$ apart.

    \item Remaining path length $d_{end}$.
\end{itemize}

\textbf{Action space}—%
The action space comprises the jerk and the change in steering rate. Because both actions are higher-order derivatives, they increase the complexity of the RL problem. However, they also reduce the sim-to-real gap, as the numerous inertia and delay effects of the real vehicle can be bridged more easily and improve smooth driving policies.

\begin{figure}[t!]
    \centering
    \includegraphics[width=\columnwidth]{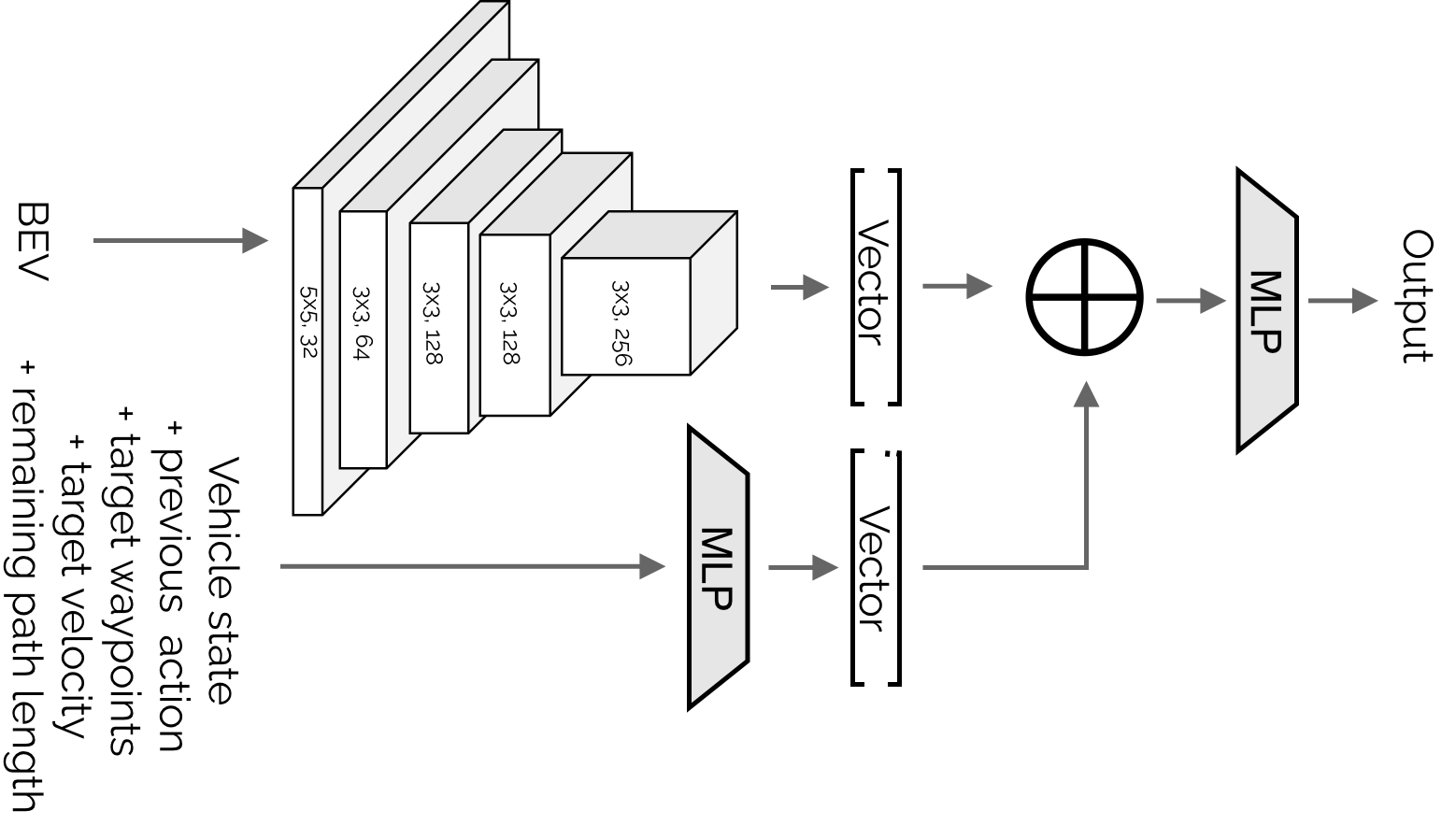}
    \caption{The policy network architecture. The BEV is processed by a CNN and subsequently flattened. The vehicle state, previous actions, target waypoints, target velocity and remaining path length are processed by an MLP.
    All resulting vectors are concatenated and fed into a final MLP to produce the output.} 
    \label{fig:network}
\end{figure}

\textbf{Network}—%
The network architecture is illustrated in \Cref{fig:network}. We use a CNN rather than a Vision Transformer because of its preferable real-time capabilities and better sample efficiency~\cite{tao2022evaluating}.
The encoder comprises five strided convolutions: one $5\times5$ followed by four $3\times3$, all with stride $2$ and $32$, $64$, $128$, $128$ and $256$ channels. Each is followed by layer normalization and ReLU.
The resulting $256\times8\times8$ feature map is flattened and projected to a $512$-dimensional embedding. The remaining $168$ state values are encoded by a two-layer MLP with $128$ units and concatenated with the BEV embedding. Separate two-layer actor and critic heads with $256$ units decode the fused $640$-dimensional feature into the action logits and the state value.

\textbf{Reward}—%
Following CaRL \cite{jaeger2025carl}, the dense reward is built around a single quantity, the progress $\Delta s$ along the target waypoints; every positive term is proportional to it. 
The progress is clipped to $(v_\mathrm{target} + \SI{0.5}{\metre\per\second})\Delta t$ and gated by a Gaussian of the lateral deviation to the target waypoints.
The remaining criteria enter additively, so violating one does not lower the amplitude of the others.

\begin{itemize}
    \item \textbf{Target velocity:} peaks at the randomly sampled target velocity and decays to zero at standstill and at \SI{1}{\metre\per\second} above the target, so that the target is approached but not exceeded.
    \item \textbf{Heading error:} peaks when the vehicle is aligned with the path and decays to zero at a heading error of $\pm\SI{0.5}{\radian}$.

\end{itemize}
Two penalties are applied at every step, independently of the progress:
\begin{itemize}
    \item \textbf{Smoothness:} the magnitudes of the acceleration and of the steering rate are penalized, which discourages aggressive oscillations and corrections.

    \item \textbf{Safe velocities:} the agent is penalized for exceeding a safety velocity, given by the minimum of three limits. The first is a specified upper limit for the centripetal acceleration. The second is a clearance limit that ramps up over an obstacle distance from $\SI{0.25}{\metre}$ to $\SI{4}{\metre}$, to avoid high velocities next to obstacles. The third is a terrain limit that interpolates between $\SI{1.5}{\metre\per\second}$ and the maximum velocity according to the fraction of \texttt{Drivable} cells among the traversable cells beneath the vehicle. The last limit implicitly enforces low velocities when driving on \texttt{Semi-drivable} ground, which comprises, for example, tall grass, where the vehicle must travel slowly and driving is riskier owing to the partially observable nature of such areas.
\end{itemize}

Finally, the agent is penalized on terminal states for colliding with non-drivable cells or moving out of bounds, and receives no reward when it comes to a standstill without having reached the goal. 
Stopping at the goal yields a positive reward that follows a Gaussian in both the remaining distance and the final heading error.

\textbf{Kinematic Restrictions}—%
To reduce sim-to-real discrepancies between the simulated and real vehicles, we conservatively restrict the kinematic and dynamic capabilities such that the capabilities of the simulated vehicle are less than or equal to those of our real vehicles.

\subsection{Training}

We train the agent with the on-policy PPO~\cite{schulman2017proximal} algorithm, building on the implementation of CaRL~\cite{jaeger2025carl}. 
We deviate from CaRL in the following points: 

\textbf{Curriculum learning}—%
We apply curriculum learning to the reward function by initially tolerating a coarser stop at the goal and by penalizing undesirable behavior, such as collisions, unsafe velocities, and non-smooth driving, less severely. 
The corresponding coefficients are ramped linearly over the first $2\,\%$ of training. This facilitates rapid learning and exploration while preventing the agent from converging to the trivial solution of remaining stationary, which may appear to be the safest policy.

\textbf{Discrete Action Head}—%
We factorize the policy into two independent categorical distributions with five bins each ($5 \times 5$) for jerk and steering-rate change.
Since our actions are integrated twice before affecting the pose, coarse quantization still yields smooth motion.
Furthermore, discretized actions allow the unambiguous recognition of the intention to remain stationary.

\textbf{Discount factor $\gamma=1$}—%
We chose $\gamma=1$ to achieve global optimization and prevent hasty agent behavior. 
This makes training more challenging and requires longer bootstrap sequences. However, the agent has an infinite amount of time to accumulate rewards, which discourages risky and overly aggressive driving behavior. 
For accurate value estimation of each state, we provide the remaining distance to the final target waypoint.

\textbf{State normalization}—%
Prior to training, we estimate the mean and standard deviation of the state vector from $50\,000$ samples collected using a random policy and use them to standardize the state inputs to approximately zero mean and unit variance.
The normalization is part of the network, so that it is self-contained and training and deployment cannot drift apart.

\subsection{Deployment}
\label{seq:deployment}

The actions output by the policy network described in \Cref{sec:modelling} are applied directly to the bicycle model during training in the MLR simulator. For deployment on the Jetson AGX Orin, the trained policy is exported to ONNX and compiled into a TensorRT engine on the target device, achieving a mean inference time of \SI{1.9}{\milli\second}.

For ground-robotics applications at low velocities, such as parking, the discrepancy between simulation and reality can often be neglected, allowing the actions to be applied directly to the real platform. However, as velocity and vehicle dynamics increase, this discrepancy can cause severe instabilities. We therefore build on the sim-to-real approach of~\cite{steinecker2025dynamics} and extend it to perception inputs represented by spatial MLRs, such as our BEVs, for which different coordinate frames can be aligned using homogeneous transformations ${}^{\text{target}}\mathbf{T}_{\text{source}}$.

We refer interested readers to~\cite{steinecker2025dynamics} for the complete approach and provide only a simplified description here. During deployment, we initialize a virtual vehicle ($V$) at the pose of the real vehicle ($R$), where the superscripts $V$ and $R$ denote the respective coordinate frames and the subscript $t$ denotes the time step. At each time step, BEVFusion generates a BEV centered on the real vehicle:
\begin{equation}
    \mathrm{BEV}_t^R
    =
    \mathrm{BEVFusion}
    \left(
        \mathrm{PointCloud}_t^R,
        \mathrm{CameraImages}_t^R
    \right).
    \label{eq:real_vehicle_bev}
\end{equation}

As the policy operates with respect to the virtual vehicle, $\mathrm{BEV}_t^R$ must be transformed into its coordinate frame. Direct transformation would introduce undefined cells near the BEV boundaries, particularly under rotational offsets. We therefore apply a circular crop to eliminate rotational artifacts and reduce its radius to account for translational artifacts:
\begin{equation}
    r_{\text{circular}}
    =
    \frac{1}{2}W_{\mathrm{BEV}}
    -
    d_{\mathrm{max}},
    \label{eq:circular_bev_radius}
\end{equation}
where $d_{\mathrm{max}}$ denotes the maximum expected distance between the real and virtual vehicles and $W_{\mathrm{BEV}}$ the BEV width.

\begin{figure}[t!]
    \centering
    \includegraphics[width=\columnwidth]{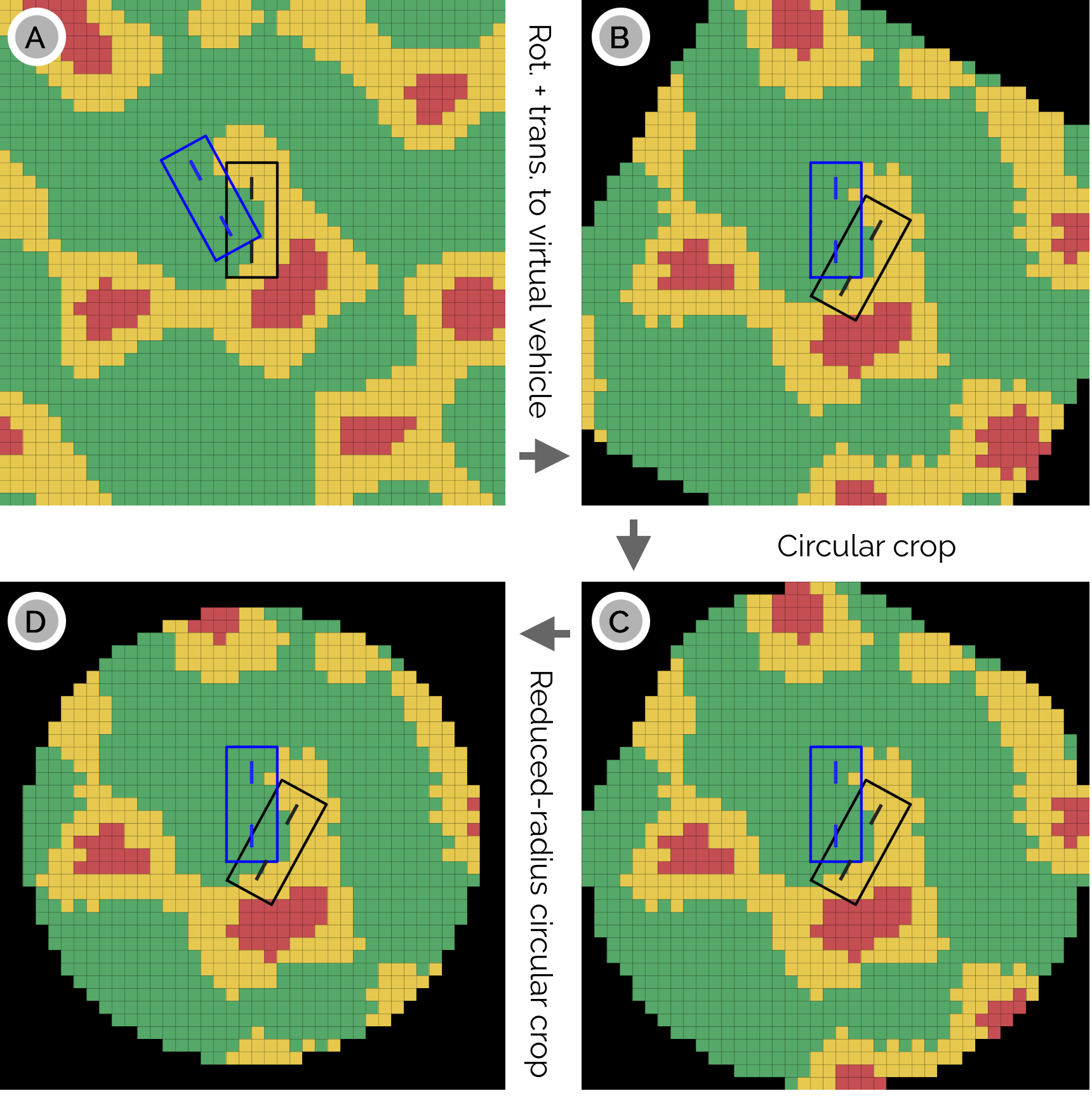}
    \caption{
    Transformation of the perception inputs of the real vehicle (black) and virtual vehicle (blue), illustrated using a simplified BEV of size $\SI{20}{\metre}\times\SI{20}{\metre}$ and exaggerated lateral and rotational offsets. \textbf{(A)} Original BEV produced by BEVFusion and centered on the real vehicle. \textbf{(B)} BEV transformed into the virtual-vehicle frame using nearest-neighbor cell assignment; cells outside the original BEV are assigned to the \textit{None} class. \textbf{(C)} Circular cropping with radius $\frac{1}{2}W_{\mathrm{BEV}}$ eliminates rotational artifacts. \textbf{(D)} Reducing the radius to $\frac{1}{2}W_{\mathrm{BEV}}-d_{\mathrm{max}}$ eliminates translational artifacts for offsets of up to $d_{\mathrm{max}}$.
    }
    \label{fig:perception_alignment}
\end{figure}

The aligned BEV is then given by
\begin{equation}
    \mathrm{BEV}_{\mathrm{circ}, t}^V
    =
    \mathrm{PerceptionTransformation}
    \left(
        \mathrm{BEV}_t^R,
        {}^V\mathbf{T}_R
    \right),
    \label{eq:perception_alignment}
\end{equation}
and the perception-transformation process is illustrated in \Cref{fig:perception_alignment}. The same crop is applied in simulation to ensure identical input dimensions and structure during training and deployment. Unlike raw sensor data, spatial MLRs such as BEVs or voxel grids can be aligned through direct translation and rotation without requiring a new sensor rendering.

The target waypoints are likewise transformed into the virtual-vehicle frame:
\begin{equation}
    \begin{bmatrix}
        x_{\mathrm{wp},i}^V \\
        y_{\mathrm{wp},i}^V \\
        1
    \end{bmatrix}
    =
    {}^V\mathbf{T}_R
    \begin{bmatrix}
        x_{\mathrm{wp},i}^R \\
        y_{\mathrm{wp},i}^R \\
        1
    \end{bmatrix},
    \qquad
    i=1,\ldots,N_{\mathrm{wp}}.
    \label{eq:waypoint_alignment}
\end{equation}

The policy network then computes the next virtual action:
\begin{equation}
    \mathbf{a}_{t}^V
    =
    \pi
    \left(
        \mathrm{BEV}_{\mathrm{circ}, t}^V,
        \mathbf{s}_{\mathrm{veh}, t}^V,
        \mathbf{a}_{t-1}^V,
        v_{\mathrm{target}, t},
        \mathbf{W}_t^V,
        d_{\mathrm{end}, t}^V
    \right).
    \label{eq:virtual_action}
\end{equation}
This action is applied to the bicycle model from the MLR simulator to obtain the next virtual state:
\begin{equation}
    \mathbf{s}_{\mathrm{veh},t+1}^V
    =
    \mathrm{BicycleModel}
    \left(
        \mathbf{s}_{\mathrm{veh},t}^V,
        \mathbf{a}_{t}^V
    \right).
    \label{eq:virtual_vehicle_state}
\end{equation}

Finally, longitudinal and lateral controllers, including a Stanley controller, align the real vehicle with the virtual trajectory to minimize their spatio-temporal discrepancy:
\begin{equation}
    \mathbf{a}_{t}^R
    =
    \mathrm{AlignmentControl}
    \left(
        [\mathbf{s}_{\mathrm{veh},t+1}^V,
         \mathbf{s}_{\mathrm{veh},t}^V,\ldots],
        \mathbf{s}_{\mathrm{veh},t}^R
    \right).
    \label{eq:control_alignment}
\end{equation}


\section{Evaluation}

\begin{figure*}[t!]
    \centering
    \includegraphics[width=1.0\textwidth]{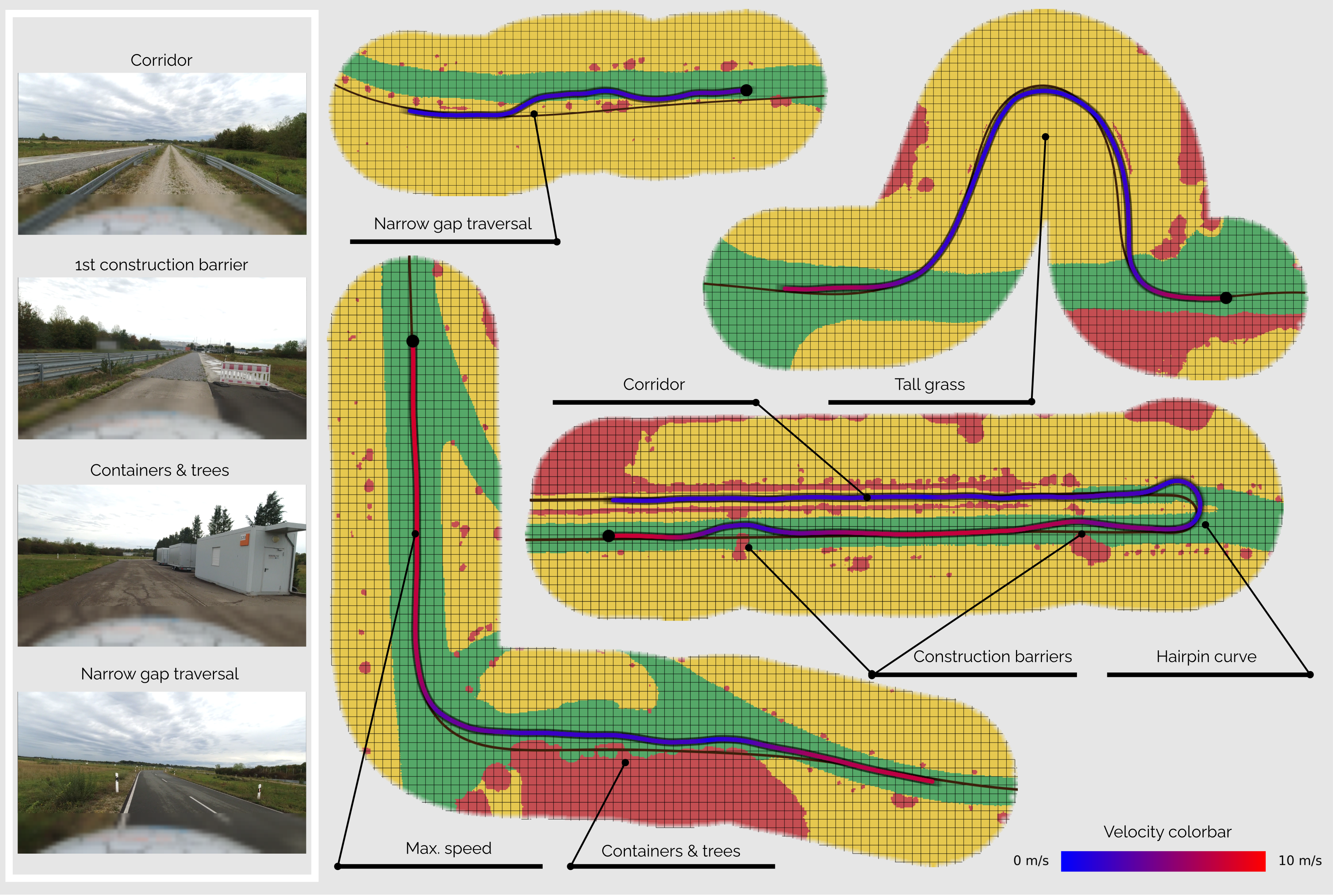}
    \caption{Selected scenarios from the real-vehicle tests. The figure shows fused BEVs, the target waypoints (brown), the ego vehicle’s position with color-coded velocity, the start of each scenario (black dot), a grid with \SI{2}{\meter} $\times$ \SI{2}{\meter} cells and selected images from the scenarios on the left. For improved visibility, we reduced the BEV radius to \SI{20}{\meter}. All BEVs are fused using the orientation of the first BEV, and newer BEVs overwrite older ones.}
    \label{fig:stitched_BEVs}
\end{figure*}

\subsection{Simulation Training}

The agent was trained in simulation using $10^9$ samples, corresponding to more than $2\times10^6$ episodes, taking approximately $68$ hours on one NVIDIA A100. 
The policy was optimised with PPO using Adam at a learning rate of $2.5\times10^{-4}$, decayed linearly to zero over training, with gradients clipped to a global norm of $0.5$. 
Advantages were computed with Generalized Advantage Estimation~\cite{schulman2015high} ($\lambda = 0.95$, $\gamma = 1.0$) and normalized over the full batch; the surrogate objective used a clipping range of $0.1$, the value loss was weighted by $0.5$ and left unclipped, and the entropy bonus was annealed linearly from $0.01$ to $0.001$. The kinematic and dynamic constraints are summarized in \Cref{tab:kinematic_constraints}.

\subsection{Deployment Experiments}

We deploy our policy on two real vehicles (A: MuCAR\mbox{-}4 and B: MuCAR\mbox{-}3) to demonstrate its applicability to different platforms. From a training perspective, the only vehicle-specific parameter is the wheelbase ($L_{\mathrm{wb}}^A$ = \SI{2.60}{\metre} and $L_{\mathrm{wb}}^B$ = \SI{2.86}{\metre}). We nevertheless train the policy using only the wheelbase of vehicle A and obtain comparable results for both vehicles.

The main differences between the runs for both vehicles are:
\begin{itemize}
    \item wheelbase $L_{wb}$: (A: 2.60\,m, B: 2.86\,m)
    \item vehicle mass and dynamics: (1.8\,t vs. 2.5\,t, different steering limits)
    \item low-level control and actuation: velocity and steering-angle loops run in each vehicle's dSPACE unit, with different steering-rate and acceleration limits
    \item BEV: vehicle-specific camera mounts and calibration
    \item weather and lighting: daylight, dry (Vehicle A) vs.\ evening, rain (Vehicle B)
\end{itemize}

\begin{figure}[t!]
    \centering
    \includegraphics[width=0.8\columnwidth]{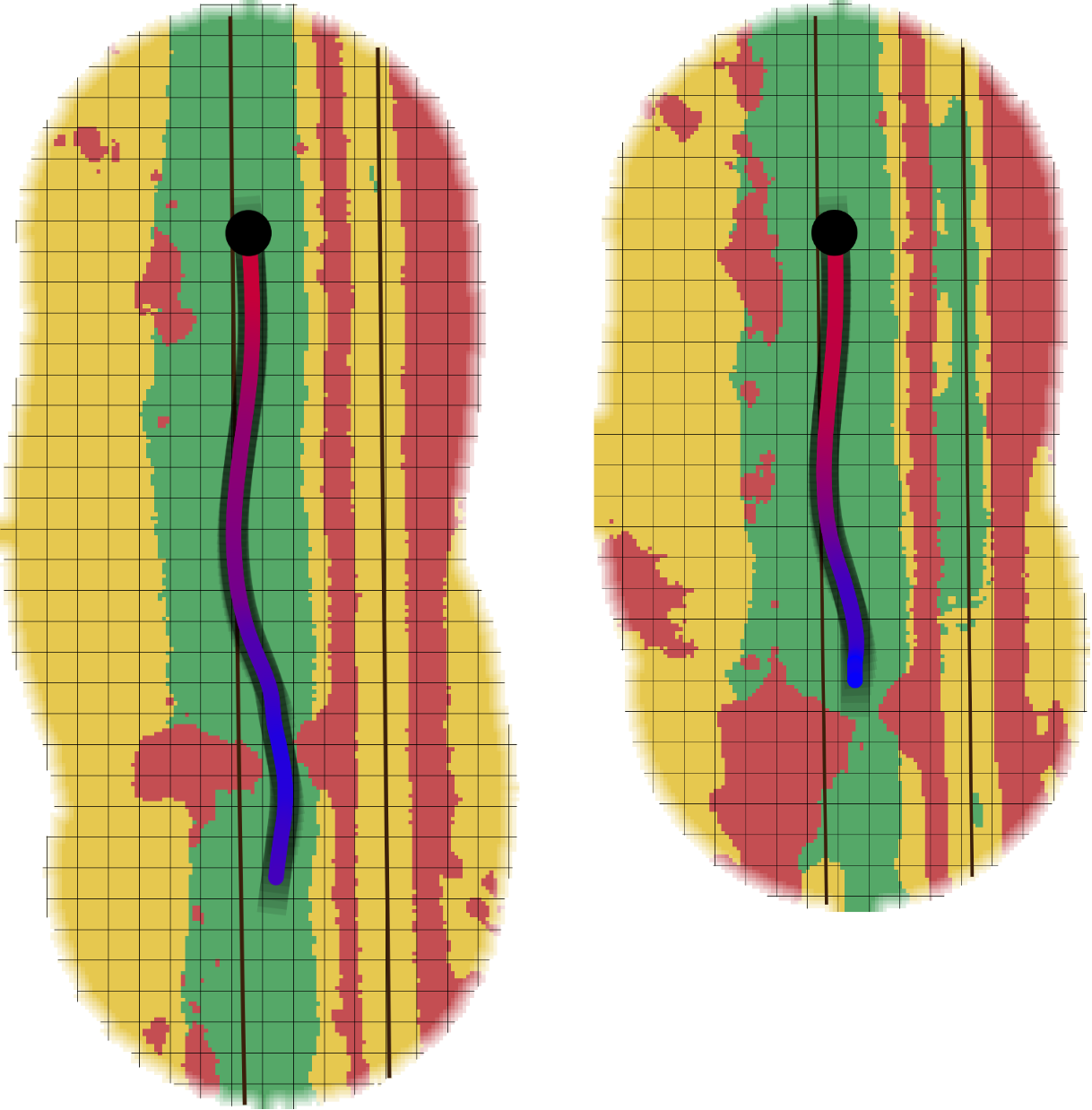}
    \caption{A comparison of the second and third rounds of vehicle B. The second round is illustrated on the left, showing that the vehicle successfully navigated through the narrow gap. In the third round, vehicle B could no longer fit through the gap and stopped. Darker conditions and rainfall strongly contributed to the differences in the appearance of the BEVs over the course of the evaluation.} 
    \label{fig:touareg_blockage}
\end{figure}

The test track is \SI{3.0}{\kilo\meter} long, and we conducted our experiments by driving the track three times in a row with each vehicle without any human intervention. 
In \Cref{fig:stitched_BEVs} we show the driving performance for the most challenging parts of the tracks for vehicle A. 
However, during the last round, vehicle B stopped due to overly conservative obstacle prediction (see \Cref{fig:touareg_blockage}), most likely because of rainfall and darker conditions. This corresponds to a total distance of \SI{17.3}{\kilo\meter}. The test track included a straight road for reaching maximum speed, large obstacles (trees, containers) and smaller obstacles (bushes, poles, construction barriers), many hairpin curves, multiple small off-road sections with tall grass or gravel, and corridors. We report aggregate performance metrics for both vehicles in \Cref{tab:vehicle_results} and provide a video of vehicle A completing all three rounds.

\begin{table}[t]
\centering
\caption{Kinematic and dynamic constraints that are sufficiently conservative for vehicles A and B.}
\label{tab:kinematic_constraints}
\begin{threeparttable}
\begin{tabular}{llrl}
\toprule
Quantity & Symbol & Value & Unit \\
\midrule
Maximum velocity
    & $v_{\max}$                         
    & $12.0$    
    & m/s \\
Minimum velocity
    & $v_{\min}$                         
    & $0.0$     
    & m/s \\
Max./Min. acceleration
    & $a_{\max},\,a_{\min}$              
    & $\pm 2.0$ 
    & m/s$^2$ \\
Max./Min. jerk
    & $j_{\max},\,j_{\min}$              
    & $\pm 2.0$ 
    & m/s$^3$ \\
Max./Min. steering angle
    & $\delta_{\max},\,\delta_{\min}$    
    & $\pm 0.44$
    & rad \\
Max./Min. steering rate
    & $\dot{\delta}_{\max},\,\dot{\delta}_{\min}$
    & $\pm 0.2$
    & rad/s \\
Max./Min. steering-rate change
    & $\ddot{\delta}_{\max},\,\ddot{\delta}_{\min}$
    & $\pm 0.2$
    & rad/s$^2$ \\
Hard centripetal-acceleration limit\tnote{a}
    & $a_{\mathrm{cp},\mathrm{hard}}$
    & $3.0$
    & m/s$^2$ \\
Soft centripetal-acceleration limit\tnote{b}
    & $a_{\mathrm{cp},\mathrm{soft}}$
    & $1.25$
    & m/s$^2$ \\
\bottomrule
\end{tabular}
\begin{tablenotes}[flushleft]
\footnotesize
\item[a] The bicycle model prevents the vehicle from exceeding this limit.
\item[b] The agent is penalized for exceeding this limit.
\end{tablenotes}
\end{threeparttable}
\end{table}

In the left scenario in \Cref{fig:stitched_BEVs}, the vehicle travels at its maximum speed (\SI{33.6}{\kilo\meter\per\hour}) without notable oscillations and reduces its velocity when taking the curve. The target waypoints then lead between the trees and containers, and the vehicle stays to the left of the path at a reduced velocity because of its proximity to obstacles. It finally resumes higher velocities after passing the obstacles. In the top scenario, the agent dodges two construction barriers. While the agent reliably reduces its speed for the first barrier, it approaches the second barrier rather hastily. Afterwards, the agent handles a 180-degree hairpin curve before entering the off-road corridor, with barriers and vegetation on either side. In the center scenario, the target waypoints leave the road, but the agent cannot leave the road directly because its path is blocked by poles and bushes. The agent initially stays on the left side of the road until it finds a gap, which it chooses to pass through to reduce the cross-track deviation from the target waypoints. In the top-right scenario, the agent leaves the road and drives onto tall grass. As specified by the reward, the agent drives at low velocities throughout the tall-grass section and accelerates again afterwards. Throughout the track, we set the target velocity to
$v_{\mathrm{target}}=\SI{30}{\kilo\metre\per\hour}$, except on the
maximum-speed straight, where we set it to
$v_{\mathrm{target}}=\SI{40}{\kilo\metre\per\hour}$.

\begin{table}[t]
    \centering
    \caption{Evaluation results for the two vehicles, where $v$ denotes velocity,
    CTD denotes cross-track deviation, $t_{tot}$ the driving time of the evaluation, and $a_{\mathrm{cp}}$ denotes centripetal acceleration.}
    \label{tab:vehicle_results}
    \small
    \setlength{\tabcolsep}{3.5pt}
    \begin{tabular}{@{}lrrrrrr@{}}
        \toprule
        
        & \shortstack{Mean $v$\\(\si{\metre\per\second})}
        & \shortstack{Max.\ $v$\\(\si{\metre\per\second})}
        & \shortstack{Mean CTD\\(\si{\metre})}
        & \shortstack{Max.\ CTD\\(\si{\metre})}
        & \shortstack{$t_{\mathrm{tot}}$\\(\si{\minute})}
        & \shortstack{Max.\ $a_{\mathrm{cp}}$\\(\si{\metre\per\second\squared})} \\
        \midrule
        A & 4.36 & 9.33 & 0.96 & 5.16 & 33.4 & 2.19 \\
        B & 4.19 & 9.02 & 1.06 & 6.16 & 31.8 & 2.19 \\
        \bottomrule
    \end{tabular}
\end{table}

\subsection{Ablations}
\textbf{Residual Gap under Identical BEVs}—We drove both vehicles on a simpler track (no obstacles, paved, \SI{1.8}{\kilo\meter}):
\begin{enumerate}
\item using the original BEVs from the deployed BEVFusion model,
\item and using identical BEVs derived for both vehicles from an HD map in OpenDRIVE format, such that both vehicles receive identical observations at the same pose.
\end{enumerate}
This reduces the remaining gap between the vehicles to hardware-related factors (wheelbase, mass, low-level control, and actuation), allowing us to determine, by comparing the two conditions, how much of the residual gap is due to differences in BEV semantics. 
The results are summarized in \Cref{tab:ablation_residual_gap}. The mean Euclidean distance is calculated from the minimum distance between the two vehicles' trajectories for each recorded pose of vehicles A and B. Additionally, we calculate the mean velocity difference between the mutually closest poses of vehicles A and B. The results indicate that BEV differences cause substantial discrepancies, whereas identical BEVs yield consistent policy behavior across vehicles, with a mean velocity difference of only \SI{0.150}{\meter\per\second}.

\begin{figure}[t!]
    \centering
    \includegraphics[width=\columnwidth]{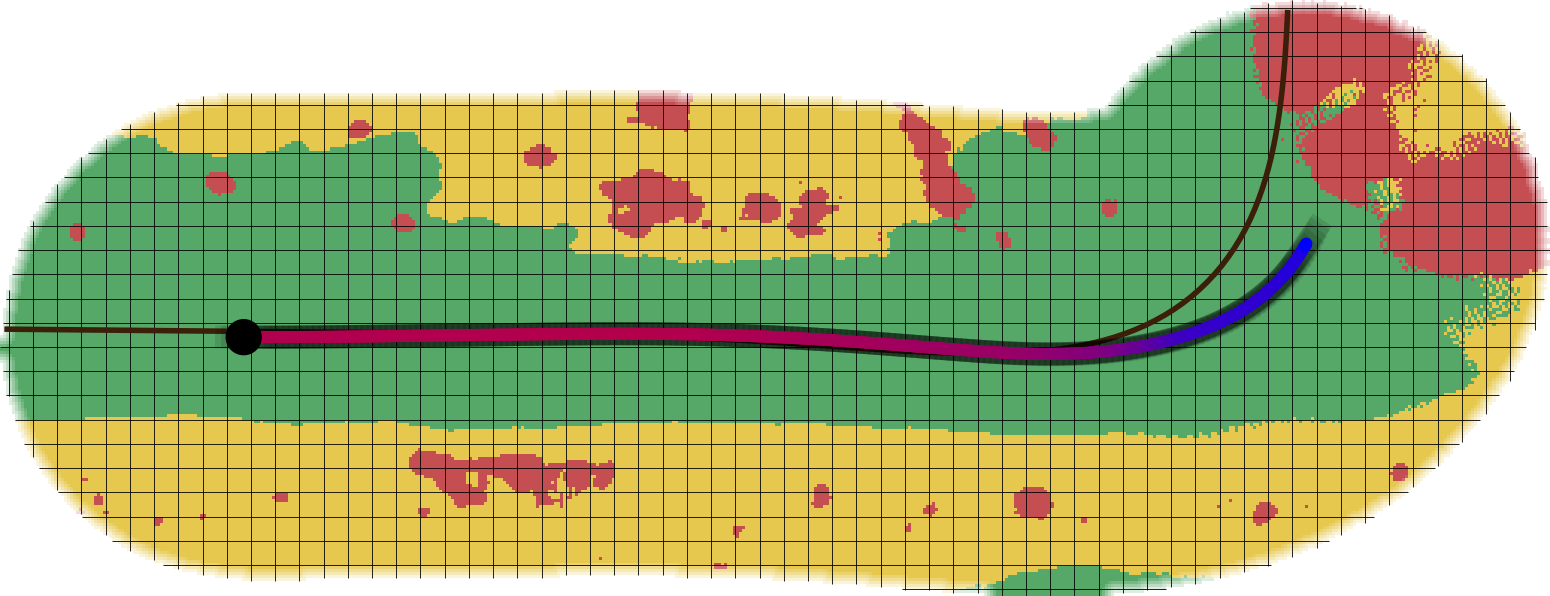}
    \caption{Control ablation with vehicle B. The scenario shows how the vehicle failed to navigate the first curve of the track, driving toward an obstacle and stopping in front of it.} 
    \label{fig:ablation_control}
\end{figure}

\textbf{Driving Directly with Control Commands}—We remove the sim-to-real mechanism described in \Cref{seq:deployment} and apply the control outputs directly by deriving acceleration and steering rate from jerk and steering rate changes. We occasionally observe strong oscillations and delayed maneuvers. This becomes apparent in \Cref{fig:ablation_control}, where the agent fails to navigate the first curve.

\begin{table}[t]
    \centering
    \caption{Ablation: Residual gap between the OpenDRIVE BEV and the original BEV.}
    \label{tab:ablation_residual_gap}
    \small
    \setlength{\tabcolsep}{8pt}
    \begin{tabular}{@{}lcc@{}}
        \toprule
        & \shortstack{Mean Eucl. distance\\(\si{\metre})}
        & \shortstack{Mean velocity diff.\\(\si{\metre\per\second})} \\
        \midrule
        OpenDRIVE BEV & 0.078 & 0.150 \\
        Original BEV  & 0.124 & 0.317 \\
        \bottomrule
    \end{tabular}
\end{table}

\section{Limitations}

As shown in the hairpin curve scenario in \Cref{fig:stitched_BEVs}, the agent does not make a human-like swing-out maneuver before the hairpin curve or dodge the second construction barrier at the same reduced velocity it uses for the first construction barrier.
Similarly, in the container and trees scenario in \Cref{fig:stitched_BEVs}, the agent does not deviate as much as it could or maintain the higher velocity while dodging.
This reflects a tradeoff in how much freedom the planner should have. The target waypoints can encode prior knowledge of drivable areas, but if the planner deviates too much from them, it can maneuver into situations where it can no longer drive, because areas initially perceived as drivable are later reclassified as semi-drivable or non-drivable. 

During training, the agent’s maximum velocity remained limited even when the desired velocity exceeded \SI{40}{\kilo\metre\per\hour}. This appears to result from the limited BEV range: the minimum braking distance is \SI{21.4}{\metre} at \SI{30}{\kilo\metre\per\hour} and \SI{28.4}{\metre} at \SI{35}{\kilo\metre\per\hour} for a jerk of \SI{-2.0}{\metre\per\second\cubed} and an acceleration of \SI{-2.0}{\metre\per\second\squared} (see \Cref{tab:kinematic_constraints}). Higher velocities therefore require greater foresight.


\section{Conclusion}

We propose MILER, an end-to-end framework that uses a semantic mid-level representation (MLR) as a common representation between simulation and deployment for unstructured autonomous driving. We extend the trajectory-alignment strategy to MLR-based perception and demonstrate zero-shot deployment in many complex, previously unseen real-world scenarios in unstructured environments. We further demonstrate the robustness of our sim-to-real strategy by deploying the same policy network on two different vehicles without human intervention. Even at velocities above \SI{30}{\kilo\metre\per\hour}, no notable oscillations occur. However, even after one billion training samples, the agent fails to discover sophisticated, human-like maneuvers, such as swinging out before entering a hairpin curve. In future work, we aim to investigate improved exploration techniques or include recorded human-driving data to enable more sophisticated driving behaviors and incorporate other dynamic agents.

\balance
\bibliographystyle{IEEEtran}
\bibliography{IEEEabrv,references}

\end{document}